\pdfoutput=1

\documentclass[11pt]{article}

\usepackage[]{EACL2023}

\usepackage{times}
\usepackage{latexsym}

\usepackage[T1]{fontenc}

\usepackage[utf8]{inputenc}

\usepackage{microtype}

\usepackage{inconsolata}
\usepackage{graphicx}
\usepackage{booktabs}
\usepackage{multirow}
\usepackage{graphics}
\usepackage{enumitem}
\usepackage{arydshln}

\title{Logic Against Bias: Textual Entailment Mitigates \\ Stereotypical Sentence Reasoning}

\author{Hongyin Luo and James Glass\\
  MIT Computer Science and Artificial Intelligence Laboratory \\
  Cambridge, MA 02139, USA\\
  \texttt{\{hyluo, glass\}@mit.edu} \\}

\begin{document}
\maketitle
\begin{abstract}
Due to their similarity-based learning objectives, pretrained sentence encoders often internalize stereotypical assumptions that reflect the social biases that exist within their training corpora. In this paper, we describe several kinds of stereotypes concerning different communities that are present in popular sentence representation models, including pretrained next sentence prediction and contrastive sentence representation models.
We compare such models to textual entailment models that learn language logic for a variety of downstream language understanding tasks.
By comparing strong pretrained models based on text similarity with textual entailment learning, we conclude that the explicit logic learning with textual entailment can significantly reduce bias and improve the recognition of social communities, without an explicit de-biasing process. The code, model, and data associated with this work are publicly available at \url{https://github.com/luohongyin/ESP.git}.

\end{abstract}

\section{Introduction}
Recent pretrained language models have achieved significant improvements on natural language understanding tasks \cite{devlin2018bert,liu2019roberta,clark2020electra,he2020deberta,brown2020language}. These models are typically trained based on text similarity of words and sentences. Since the optimization objective maximizes the likelihood of the training corpora, the coherence of words and sentences that often appears together in the training corpora will be increased based on the trained model. However, since the training corpora are generated by humans, they can contain a large amount of social bias and stereotypes, including those concerning gender, race, and religion \cite{nadeem2020stereoset,stanczak2021survey,kiritchenko2018examining}.
\begin{figure}[t]
\centering
\includegraphics[width=.8\columnwidth]{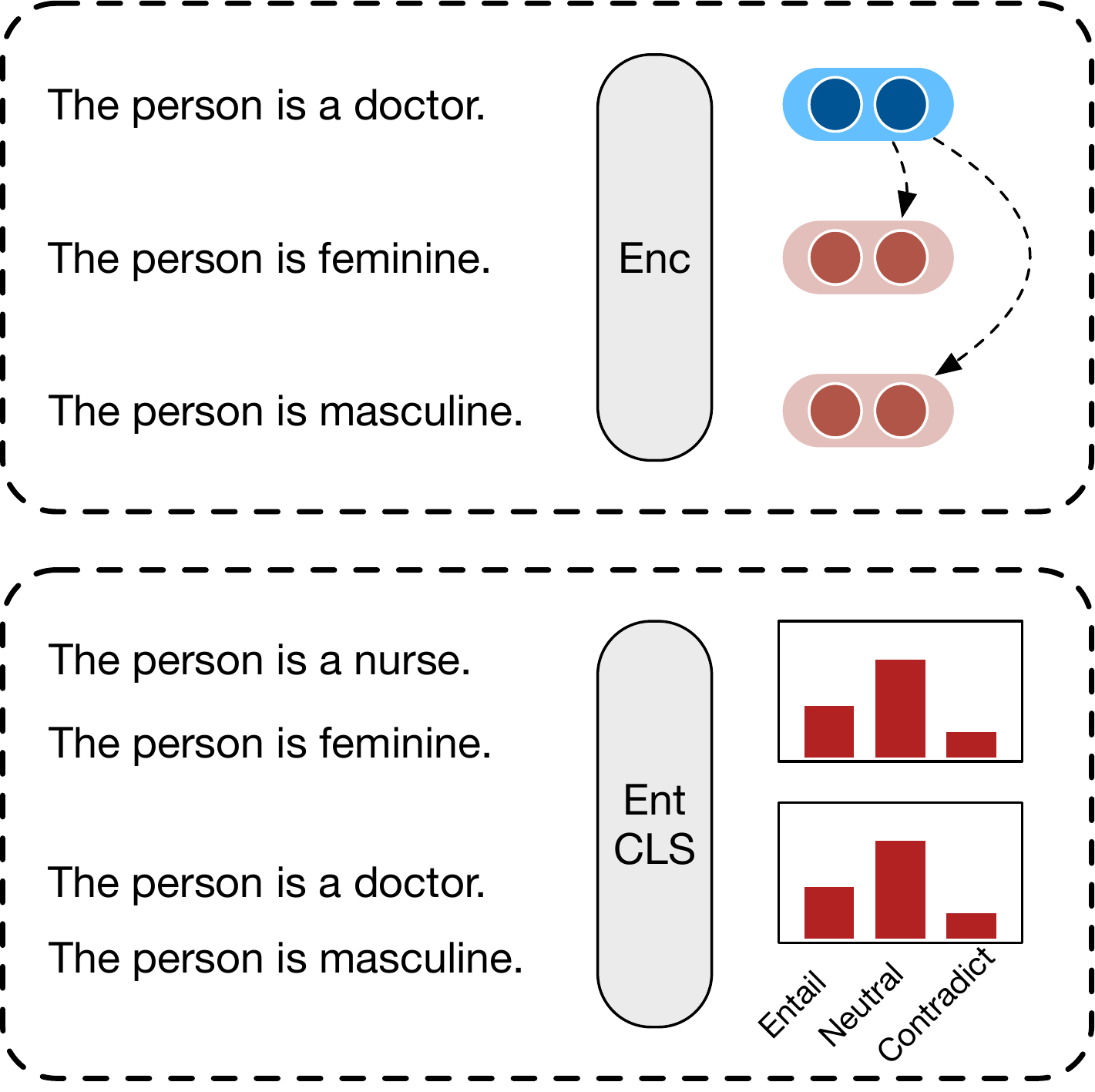}
\caption{Mitigating stereotypical sentence reasoning bias with textual entailment models. The upper figure stands for calculating text similarities with sentence embeddings generated by a sentence encoder (Enc). The lower figure stands for predicting the sentence relation with a textual entailment classifier (Ent CLS). Both sentence pairs are predicted neutral by the classifier.}
\label{fig:cover}
\end{figure}

In contrast, learning by textual entailment \cite{dagan2005pascal,N18-1101} focuses more on logic than semantic similarity. According to \citet{dagan2005pascal}, textual entailment is not necessarily strict logical entailment. Instead, textual entailment stands for the case where the premise is true so that the hypothesis is \textit{likely} to be true. Contradiction means that when the premise is true, the hypothesis is \textit{likely} to be false. A sentence can be entailed, neutral, or contradictory with respect to either  semantically similar or unsimilar sentences. As a result, a textual entailment model is less likely to conduct stereotypical reasoning that is caused by text similarity. As illustrated in Figure \ref{fig:cover}, a sentence encoder model can generate sentence representations that reflect the bias in the pretraining corpora via text similarity calculations. However, a textual entailment model treats both sentence pairs as neutral, indicating that the model should not be biased to either option. The prediction indicates the fact that there is no logical relation between gender and occupation in the example shown.

Besides gender, we also investigate different types of stereotypical sentence reasoning of language models, including race, religion, profession, and emotion using StereoSet \cite{nadeem2020stereoset}, profession and gender terms in \cite{lu2020gender}, and emotion terms in \cite{kiritchenko2018examining}. We make the following contributions in this work:

\noindent \textbf{1. Bias in sentence representations.} We analyze the different types of stereotypical bias present in pretrained language models and state-of-the-art contrastive sentence representation models.

\noindent \textbf{2. Textual entailment debiases.} 
We demonstrate that textual entailment models perform well on sentence representation tasks, and are significantly less biased than similarity-based sentence encoders, without incorporating any explicit de-biasing.

\noindent \textbf{3. Similarity causes bias, logic leads to fairness.} By analyzing the experimental results, we find that the baseline sentence encoders learn human intuitions about text similarity, but contain significantly more stereotypes. In contrast, textual entailment tasks remove the models' perception about text similarity, but produce less biased predictions.

\section{Related Work}
Recent advances in language modeling has followed the strategy of learning large-scale models on large-scale unannotated corpora with self-supervised learning, including masked word and next sentence prediction \cite{devlin2018bert,liu2019roberta,he2020deberta}, wrong word detection \cite{clark2020electra}, and left-to-right language generation \cite{brown2020language,raffel2020exploring}. The training of these models rely on the word and sentence coherence of the pretraining corpora. Word-level language models are the foundation of sentence-level language encoders, including sentenceBERT \cite{reimers2019sentence}, SimCSE \cite{gao-etal-2021-simcse}, and DiffCSE \cite{chuang-etal-2022-diffcse}, that were proposed for generating sentence embeddings with better representation abilities.

Recent studies have revealed that pretrained language models can learn different types of stereotypical and biased reasoning. \citet{recasens2013linguistic} investigated biased languages using Wikipedia texts. \citet{lu2020gender} surveyed stereotypical reasoning in word-level language prediction and co-reference resolution. \citet{kiritchenko2018examining} probed language models with the sentiment analysis task and measured the different model behaviors against different social groups. Stereotypical reasoning against race, gender, profession, and religion were also evaluated on recent masked language models and sentence encoders in \citet{nangia2020crows} and \citet{nadeem2020stereoset}.

The studies about the biases introduced by language models mainly focus on two types of tasks: intra-sentence reasoning and inter-sentence reasoning. Intra-sentence, or word-level, reasoning represents word and co-reference selection in a single sentence, which reveals the bias within word and context representations \cite{bao2019transfer,bartl2020unmasking,bolukbasi2016man,bordia2019identifying,cao2019toward,chaloner2019measuring,manzini2019black,caliskan2017semantics}. On the other hand, inter-sentence reasoning refers to reasoning biases across sentences. More specifically, a set of given sentences may not have any logical relationship, but a similarity-based language model may be biased towards linking a subset of the sentences, reflecting the coherence bias of the pretraining corpora \cite{may2019measuring,kiritchenko2018examining,nadeem2020stereoset}. Recent studies have also investigated the social bias under multi-lingual settings \cite{costa2019gebiotoolkit,elaraby2018gender,font2019equalizing}.

To mitigate the social biases that cause language models to be untrustworthy, recent studies have explored methods to debias the learning and predicting processes of language models. Typical debiasing methods include counterfactual data augmentation \cite{zmigrod2019counterfactual,dinan2019queens,webster2020measuring,barikeri2021redditbias}, dropout regularization \cite{webster2020measuring}, self-debias \cite{schick2021self}, sentence embedding debias \cite{liang2020towards}, and iterative nullspace projection \cite{ravfogel2020null}.

Besides the regular similarity-based pretraining method applied by most language models, some sentence encoding models also employ natural language inference (NLI) corpora to learn textual entailment~\cite{bowman2015large,N18-1101}. Superivised SimCSE \cite{gao-etal-2021-simcse} and SentenceBERT \cite{reimers2019sentence} use entailment data as a part of the pretraining corpora, while other studies apply entailment models to handle downstream tasks, including fact-checking \cite{thorne2018automated}, relation extraction \cite{obamuyide2018zero}, and text classification \cite{yin-etal-2019-benchmarking}. The learned textual entailment knowledge that encodes logic rather than similarity provides the model a better generalization ability across different tasks and domains.

\section{Method}
\subsection{Measuring Stereotypical Reasoning}
In this work, we use data from three different sources to measure the stereotypical biases of sentence encoders. We use the following corpora and corresponding data construction strategies:

\noindent \textbf{StereoSet.} The StereoSet corpus \cite{nadeem2020stereoset} contains both intra- and inter-sentence tasks for evaluating stereotypical reasoning, including gender, race, religion, and profession. Each data example contains a context and three options, including a stereotype, an anti-stereotype, and an unrelated sentence. A model is required to score each option and pick one. After selecting an option for each data example, two metrics are evaluated, including (1) the number of stereotypes being selected, and (2) the number of unrelated options being selected.

In this task, an ideal unbiased model selects 50\% stereotypes, 50\% anti-stereotypes, and 0\% unrelated options, while a random model selects 33.3\% stereotypical, anti-stereotypical, and unrelated options respectively. We used the idealized Context Association Test (\textit{iCAT}) score (\%) to jointly assess the quality and fairness of the sentence encoders.
\begin{equation}
\textit{iCAT} = lms \cdot \frac{min(ss, 100 - ss)}{50}
\end{equation}
where $lms$ (language model score) stands for the percentage that the model selects a related option, and $ss$ (stereotype score) stands for the percentage that the model selects a stereotypical option. The \textit{iCAT} score highlights the models that tend to select related options with no preference as to stereotypical and anti-stereotypical options.

\noindent \textbf{Gender Profession \& Emotion Test.} We apply the gender and profession vocabulary sets from \citet{lu2020gender} and the sentiment vocabulary set from \citet{kiritchenko2018examining}. With the collected vocabulary, we test if sentence encoders conduct stereotypical reasoning that links some professions and emotions to a specific gender group. We also use an \textit{iCAT} score to measure the fairness, which is calculated using different metrics
\begin{equation}
\textit{iCAT} = grs \cdot \frac{min(gbs, 100 - gbs)}{50}
\end{equation}
where $grs$ (gender recognition score) stands for the percentage that a model correctly predicts the gender of a gender-indicating noun, and $gbs$ (gender bias score) stands for the percentage that a model links a profession or emotion to the man gender. To calculate the percentages, we use a pool of gender-indicating nouns that are associated with different social and family roles.
\begin{table*}[t]
\tiny
\centering
\begin{tabular}{@{}llllll@{}}
\toprule
\multicolumn{1}{c}{\textbf{Dataset}}                 & \multicolumn{1}{c}{\textbf{Task}} & \multicolumn{1}{c}{\textbf{Context}} & \multicolumn{1}{c}{\textbf{Options}}        & \multicolumn{1}{c}{\textbf{Metric}} & \multicolumn{1}{c}{\textbf{Ideal Score}} \\ \midrule
StereoSet                                         & Related option retrieval        & [premise claim]                              & [related \& unrelated claims]                & Language model score                & 100\%                                    \\
StereoSet                                         & Stereotype retrieval              & [premise claim]                              & [stereotypical \& anti-stereotypical claims] & Stereotype score                    & 50\%                                     \\
Gender \& profession             & Gender recognition                & {[}masculine/feminine role{]}           & The person is masculine/feminine                & Gender recognition score            & 100\%                                    \\
Gender \& profession             & Profession bias                   & The person is a {[}pro{]}            & {[}masculine/feminine role{]}     & Gender bias score                   & 50\%                                     \\
Emotion Vocab & Emotion bias                      & The person feels {[}emo{]}           & {[}masculine/feminine role{]}     & Gender bias score                   & 50\%                                     \\ \bottomrule
\end{tabular}
\caption{The summary of data, tasks, prompts, metrics, and the scores of an ideal model that will be applied for evaluation in this work. Gender \& profession stands for the corresponding vocabulary sets in \citet{lu2020gender}, and Emotion Vocab stands for the emotion vocabulary set in \citet{kiritchenko2018examining}.}
\label{tab:tasks}
\end{table*}

\subsection{Data and Prompts}
We formulate each context association test as a context sentence and three optional sentences. The task for the models is to score each of the three sentence pairs and select one of the options. We introduce how we construct the test sets and prompts respectively as follows:

\noindent \textbf{StereoSet Inter-sentence.} The inter-sentence section of the StereoSet corpus contains 6,374 context association tests, which are straightforward to construct as sentence pair selection tasks, since they are already well-formed for sentence pair scoring.

\noindent \textbf{StereoSet Intra-sentence.} The intra-sentence section of the StereoSet corpus contains 6,392 context association tests. Unlike the inter-sentence tests, intra-sentence tests were originally designed for masked language models to select a good word to fill in the \textit{BLANK} in the context sentence. For example, the context sentence of a test is ``\textit{Girls tend to be more BLANK than boys}'', and the option words are ``soft'', ``determined'', and ``fish'' (less related). To conduct sentence-level reasoning, we replace the \textit{BLANK} in the context sentence with ``what'' and the candidate words. As a result, a sentence encoder is required to represent the following sentences, ``\textit{Girls tend to be more what than boys}'' and ``\textit{Girls tend to be more soft than boys}'', etc.

\noindent \textbf{Gender-indicating terms.} We collect 71 pairs, or 142 binary gender-indicating terms about social and family roles from \citet{lu2020gender}, for example, uncle and aunt. 71 of them are masculine and the other 71 are feminine. For each term, for example \textit{aunt}, we construct a prompt ``the person is a(n) \textit{aunt}''. We evaluate if a model successfully reasons ``\textit{the person is a(n) aunt}'' $\rightarrow$ ``\textit{the person is feminine.}'' The motivation for this gender recognition test is two-fold. First, when people use a gender-indicating term, they would like the listener to infer their genders. Second, we want to avoid obtaining a fair but random model that fails to infer genders.

\noindent \textbf{Professions and emotions.} We collect 65 occupation names from \citet{lu2020gender}, 20 emotion state terms, and 20 emotional situation terms from \citet{kiritchenko2018examining}. For an occupation term \textit{PRO}, we construct a prompt ``\textit{The person is a PRO}''; for an emotion state term \textit{ES}, we construct a prompt ``\textit{The person feels ES}''; and for an emotion situation term \textit{ESIT}, we construct a prompt ``\textit{The person told us about the ESIT event.}'' We evaluate whether a model tends to link the construct profession and emotion prompts to one of the genders or not.

A summary of the data, tasks, prompts, metrics, and scores of an ideal model is shown in Table \ref{tab:tasks}. We define an ``ideal model'' as a fair and perfectly understanding model.

\subsection{Textual Entailment}
\noindent \textbf{Training.} We train the textual entailment models with the MultiNLI corpus \cite{N18-1101}. In MultiNLI, each data example contains a premise and a hypothesis, and the task is to predict if the hypothesis is likely to be true or false given the premise. Each sentence pair is classified into three classes: entailed, neutral, and contradictory. For a premise \textit{p} and a hypothesis \textit{h}, we construct the following supposition 
for the entailment model,
\begin{center}
\textit{h} is entailed by \textit{p}.
\end{center}
The classifier model is trained to output \textit{true}, \textit{false}, and \textit{neutral} for each input supposition, and the entailment relations of each sentence pair can be directly inferred from the truth value of the corresponding prompt. In this work, we train entailment classifiers based on BERT \cite{devlin2018bert}, RoBERTa \cite{liu2019roberta}, and DeBERTa \cite{he2020deberta}.

\noindent \textbf{Evaluation.} Standard sentence reasoning methods are based on the inner product of the embeddings of two sentences. With the textual entailment models, we can calculate three scores for each sentence pair, including entail, neutral, and contradictory scores. With these scores, we can calculate a prediction about the logical relation between two sentences. In summary, we have two strategies to score sentence pairs: 1. continuous sentence pair scoring with entail, neutral, or contradiction scores, and 2. discrete scoring using entailment predictions (entail = 0, neutral = 1, and contradictory = 2). Given a context, we prefer an option with a higher entailment score, lower contradictory score, and smaller entailment labels.

For the continuous scoring strategy, we calculate the language model score with the number of tests where the stereotype or anti-stereotype option score is higher than the unrelated option, and calculate the stereotype score with the number tests where stereotype option score is higher than anti-stereotype option. For the discrete scoring strategy where we assign each option an entailment label, the language score is calculated with the number of tests where the unrelated option is predicted to be less entailed than the stereotype or anti-stereotype. The stereotype score is calculated with the number of tests where the label $\{0, 1, 2\}$ of the stereotype option is lower then the anti-stereotype.

\section{Experiments}
\subsection{Language Understanding}
To ensure that the fairness of the entailment-based language model does not come from a lack of language understanding ability, we first show the zero-shot adaptation performance of the entailment-based language models. On the MNLI-mismatch task, The RoBERTa model achieves 89.0\% accuracy, and the DeBERTa model achieves 83.4\%. We compare different language models on other tasks in the GLUE benchmark \cite{wang2018glue}, including QNLI, QQP, RTE, and SST2 tasks. For each task, we construct suppositions for classification according to the corresponding task description as shown in Table \ref{tab:sup}.
\begin{table}[h]
\small
\centering
\begin{tabular}{@{}lll@{}}
\toprule
\multicolumn{1}{c}{\textbf{Task}} & \multicolumn{1}{c}{\textbf{Inputs}} & \multicolumn{1}{c}{\textbf{Supposition}} \\ \midrule
MNLI  & \{\textit{p}, \textit{h}\}    & \textit{h} is entailed by \textit{p}.   \\
RTE   & \{\textit{p}, \textit{h}\}    & \textit{h} is entailed by \textit{p}.   \\
QNLI  & \{\textit{p}, \textit{q}\}    & The answer to \textit{q} is entailed by \textit{p}.    \\
QQP   & \{\textit{x}, \textit{y}\}      & \textit{x}'s answer is entailed by \textit{y}'s answer.  \\
SST2  & \{\textit{r}\}       & The movie is good is entailed by \textit{r}. \\
\bottomrule
\end{tabular}
\caption{The suppositions constructed based on the definitions of different GLUE tasks \cite{wang2018glue}.}
\label{tab:sup}
\end{table}

We compare the zero-shot adaptation performance of our entailment-based supposition (ESP) language models with weakly supervised baseline models of different scales as follows:

\noindent \textbf{Few-shot 350M models.} We compare our entailment-based models with LM-BFF \cite{gao2020making} and UPT \cite{wang2022towards} models. Both baseline models are based on \texttt{RoBERTa-large} that contains 350M parameters with 32 human-annotated training samples.

\noindent \textbf{Few-shot 137B models.} We also compare the entailment-based models with large-scale language models (LLMs), LaMDA \cite{thoppilan2022lamda} and FLAN \cite{wei2021finetuned} containing 137B parameters, which are about 400 times larger than the entailment-based models. The LLMs are adapted to the tasks with 4 to 8 training samples.
\begin{table}[]
\centering
\begin{tabular}{@{}lccccc@{}}
\toprule
Method   & QNLI          & QQP           & RTE           & SST2          & Avg.          \\ \midrule
\multicolumn{6}{c}{Few-shot 350M models}                                                 \\ \hdashline[1.5pt/2pt]
LM-BFF   & 69.2          & 69.8          & 83.9          & 90.3          & 78.3          \\
UPT      & 70.1          & 72.1          & 68.9          & 92.9          & 76.0          \\ \midrule
\multicolumn{6}{c}{Few-shot Large-scale 137B models}                                     \\ \hdashline[1.5pt/2pt]
LaMDA    & 55.7          & 58.9          & 70.8          & 92.3          & 69.4          \\
FLAN     & 63.3          & 75.9          & \textbf{84.5} & \textbf{94.6} & 79.6          \\ \midrule
\multicolumn{6}{c}{Zero-shot entailment-based 350M model}                                \\ \hdashline[1.5pt/2pt]
RoBERTa  & 71.5          & 78.6          & 81.2          & 87.7          & 79.8          \\
DeBERTa  & \textbf{77.3} & \textbf{79.9} & \textbf{84.5} & 90.1          & \textbf{82.9} \\ \bottomrule
\end{tabular}
\caption{The performance of zero-shot entailment-based models and strong few-shot supervised baselines.}
\label{tab:nlu}
\end{table}

The results are shown in Table \ref{tab:nlu}. We found that overall, both RoBERTa and DeBERTa-based entailment models outperform all baselines, without using any task-specific training data. This proves the computation and data efficiency of entailment-based language models.

\subsection{Fairness}
We evaluate pretrained language models, supervised/unsupervised SimCSE \cite{gao-etal-2021-simcse}, and entailment models based on BERT, RoBERTa, and DeBERTa. The overall experiment results are shown in Table \ref{tab:results}.
\begin{table*}[]
\centering
\resizebox{\textwidth}{!}{
\begin{tabular}{lllllllllllllll}
\hline
\multirow{2}{*}{\textbf{Model}}   & \multicolumn{3}{c}{\textbf{StereoSet-Intra}}                                & \multicolumn{3}{c}{\textbf{StereoSet-Inter}}                                & \multicolumn{2}{c}{\textbf{Gender recog.}}         & \multicolumn{3}{c}{\textbf{Profession}}                                       & \multicolumn{3}{c}{\textbf{Emotion}}                                          \\ \cline{2-15} 
                                  & \multicolumn{1}{c}{LMS} & \multicolumn{1}{c}{FS} & \multicolumn{1}{c}{iCAT} & \multicolumn{1}{c}{LMS} & \multicolumn{1}{c}{FS} & \multicolumn{1}{c}{iCAT} & \multicolumn{1}{c}{Mean} & \multicolumn{1}{c}{Std} & \multicolumn{1}{c}{Mean} & \multicolumn{1}{c}{Std} & \multicolumn{1}{c}{iCAT} & \multicolumn{1}{c}{Mean} & \multicolumn{1}{c}{Std} & \multicolumn{1}{c}{iCAT} \\ \hline
BERT                              & 78.52                   & 89.90                  & 70.58                    & 79.02                   & 93.44                  & 73.84                    & 64.08                    & 24.90                   & 91.27                    & 6.84                    & 58.49                    & 94.51                    & 4.03                    & 60.56                    \\
\multicolumn{1}{r}{-SimCSE-unsup} & 89.46                   & 83.38                  & 74.59                    & 90.40                   & 81.36                  & 73.55                    & 85.92                    & 11.95                   & 68.78                    & 21.90                   & 59.10                    & 69.01                    & 22.18                   & 59.29                    \\
\multicolumn{1}{r}{-SimCSE-sup}   & 79.83                   & 74.82                  & 59.73                    & 91.61                   & 80.68                  & 73.90                    & 97.18                    & 2.00                    & 30.51                    & 24.21                   & 29.65                    & 40.49                    & 20.80                   & 39.35                    \\
RoBERTa                           & 32.18                   & 96.78                  & 16.14                    & 57.22                   & 96.04                  & 45.95                    & 57.04                    & 12.95                   & 72.68                    & 15.94                   & 41.46                    & 50.70                    & 9.70                    & 28.92                    \\
\multicolumn{1}{r}{-SimCSE-unsup} & 59.01                   & 82.72                  & 48.82                    & 90.10                   & 81.86                  & 73.76                    & 88.03                    & 10.96                   & 55.90                    & 25.33                   & 49.21                    & 67.54                    & 17.26                   & 59.46                    \\
\multicolumn{1}{r}{-SimCSE-sup}   & 64.24                   & 75.34                  & 48.40                    & 95.14                   & 80.32                  & 76.42                    & 99.30                    & 0.10                    & 42.90                    & 27.75                   & 42.60                    & 76.69                    & 4.60                    & 76.15                    \\
DeBERTa                           & 76.24                   & \textbf{99.68}         & 76.00                    & 68.90                   & 94.20                  & 64.91                    & 53.52                    & 23.91                   & 73.54                    & 13.63                   & 39.56                    & 60.21                    & 13.78                   & 32.22                    \\ \hline
BERT-Ent-Score                    & 88.95                   & 87.54                  & 77.88                    & 88.31                   & \textbf{96.96}         & 85.62                    & \textbf{100.00}          & \textbf{0.00}           & 68.56                    & 20.68                   & 68.56                    & 72.89                    & 5.72                    & 48.20                    \\
RoBERTa-Ent-Score                 & 91.77                   & 78.48                  & 72.02                    & 96.06                   & 92.16                  & 88.53                    & 99.30                    & 0.10                    & 87.54                    & 8.70                    & 86.93                    & 79.15                    & 19.98                   & 78.60                    \\
DeBERTa-Ent-Score                 & 92.88                   & 89.24                  & 82.88                    & 97.44                   & 90.96                  & 88.64                    & \textbf{100.00}          & \textbf{0.00}           & 80.56                    & \textbf{0.63}           & 80.56                    & 81.48                    & 2.68                    & 81.48                    \\
BERT-Ent-Pred                     & 90.79                   & \textbf{95.82}         & 86.99                    & 98.26                   & 96.90                  & \textbf{95.22}           & 75.00                    & 0.34                    & \textbf{98.35}           & 1.94                    & 73.76                    & 94.96                    & 3.59                    & 71.22                    \\
RoBERTa-Ent-Pred                  & \textbf{95.34}          & 92.04                  & 87.75                    & 99.25                   & 94.42                  & 93.70                    & 88.73                    & 8.96                    & 95.80                    & 4.20                    & 85.00                    & \textbf{98.77}           & \textbf{1.32}           & 87.64                    \\
DeBERTa-Ent-Pred                  & 95.31                   & 95.66                  & \textbf{91.16}           & \textbf{99.42}          & 94.04                  & 93.49                    & 97.53                    & 1.49                    & 97.51                    & 0.88                    & \textbf{95.10}           & 95.77                    & 4.13                    & \textbf{93.40}           \\ \hline
\end{tabular}
}
\caption{Performance of pretrained language models and textual entailment models on StereoSet, gender recognition (rec.), profession, and emotion tests. LMS stands for language model score, FS stands for fairness score, and iCAT stands for ideal context association test score. NSP stands for next sentence prediction. The profession and emotion iCAT scores are calculated by multiplying the gender recognition score and the corresponding fairness scores. All scores are in percentage (\%).}
\label{tab:results}
\end{table*}

\noindent \textbf{StereoSet-Intrasentence.} In Table \ref{tab:results}, we use the fairness score (FS) to assess the bias of the models. We have $FS = \frac{min(ss, 1 - ss)}{0.5}$, where $ss$ stands for the stereotype score defined in \cite{nadeem2020stereoset}. All baselines are sentence reasoning models pretrained with the next sentence prediction (NSP) task. We noticed that stronger sentence encoders can lead to more biased reasoning results. For BERT-based models, the unsupervised SimCSE model achieves a much higher language model score than the BERT-NSP model, outperforming by over 10\%. The supervised SimCSE also marginally outperforms the baseline model. However, both SimCSE models are more biased. The fair score of the supervised SimCSE is 15\% lower than the baseline BERT model. Because of the high sentence retrieval performance, the unsupervised SimCSE model achieves the best iCAT score, outperforming the pretrained BERT model by 4\%.

The result remains the same for RoBERTa-based models. Both supervised and unsupervised SimCSE models significantly outperform the pretrained model, by 27\% and 32\%, respectively. As with the BERT-based models, RoBERTa SimCSE models are also more biased. According to the low language modeling score, the baseline RoBERTa pertrained model is almost random. As a result, the fairness score is as high as 96\%. The SimCSE models achieve higher iCAT scores mainly because of the improvement on the language model score. We found that the DeBERTa model achieves the highest iCAT score among all NSP models. It achieves a very high fairness score (99.68\%), but a relatively low language model score of 76.24\%. As a result, the iCAT score of DeBERTa is only marginally higher than the BERT-based unsupervised SimCSE model, which achieves a 89.46\% language model score.

The entailment models achieve the best iCAT score, and both entailment scoring strategies outperform baseline sentence embedding models. Comparing with the best BERT, RoBERTa, and DeBERTa based baselines, the corresponding discrete entailment model achieved a 12.5\%, 39\%, and 25\% improvement in iCAT score. We observed that the discrete scoring models are generally better than the continuous scoring method. Although the continuous scoring method has certain biases, a discrete model can prevent biased prediction. For example, although the entailment score of option \textit{a} is higher than option \textit{b}, both options can be both classified into the neutral category.

\noindent \textbf{StereoSet-Intersentence.} In general, the Intersentence task had similar trends as the Intrasentence task.  The performance of the pretrained baseline models perform much better than the intrasentence tasks since the options are more diverse, making it easier for the models to identify the more related options. The difference within the baseline models are that the supervised SimCSE models perform better than the unsupervised sentence embedding models. 

The entailment models are also significantly better than all the baseline models. All discrete scoring models achieve higher than 99\% language modeling scores, and the fairness scores are all higher than 94\%. The iCAT scores of the discrete entailment models are at least 93.4\%, outperforming the best baseline model, supervised SimCSE with RoBERTa by 18\%. On the other hand, the continuous entailment models also outperform the best SimCSE model by at least 9\% in iCAT score. We also note that the discrete entailment models outperform the continuous models by a significant margin because the labels prevent a large amount of stereotypical reasoning.

\noindent \textbf{Gender recognition.} We evaluated the models' ability to recognize the gender of binary gender-indicating nouns, for example, (\textit{uncle}, \textit{aunt}) and (\textit{brother}, \textit{sister}). We use the set of 71 pairs, 142 gender-indicating nouns from \citet{lu2020gender}. The RoBERTa-based, supervised SimCSE model achieves high gender recognition accuracy (as high as 99\%), while the performance of the pretrained DeBERTa model is close to random at around 50\%. We found that the supervised SimCSE models are significantly better than other baseline models on this task.

On the other hand, we found that the continuous entailment scoring strategy achieves very high gender recognition performance. All three models achieve an accuracy higher than 99\% with very low standard deviations. In contrast to the previous tasks, the discrete scoring models have decreased performance. We hypothesize that this is because the continuous models are good enough, but the discrete model score blurs the selective bias, which is needed in this task since we need diverse predictions. Despite this fact, the DeBERTa based discrete model still achieves high gender recognition accuracy (97\%).

\noindent \textbf{Profession bias test.} We use a vocabulary set from \citet{lu2020gender} consisting of 65 profession nouns which are expected be gender-neutral, but possibly being affected by stereotypes. For the baseline models, we found that the stronger sentence representation models, supervised and unsupervised SimCSE, are significantly more biased than pretrained language models. Since the SimCSE models learns better sentence embeddings based on text similarity, they perform better at gender recognition, but retain more stereotypes in the pretraining corpora. Combined with the high gender recognition performance, the unsupervised BERT SimCSE model achieves the best iCAT score among all baseline models.

\begin{figure*}[t]
\centering
\includegraphics[width=\textwidth]{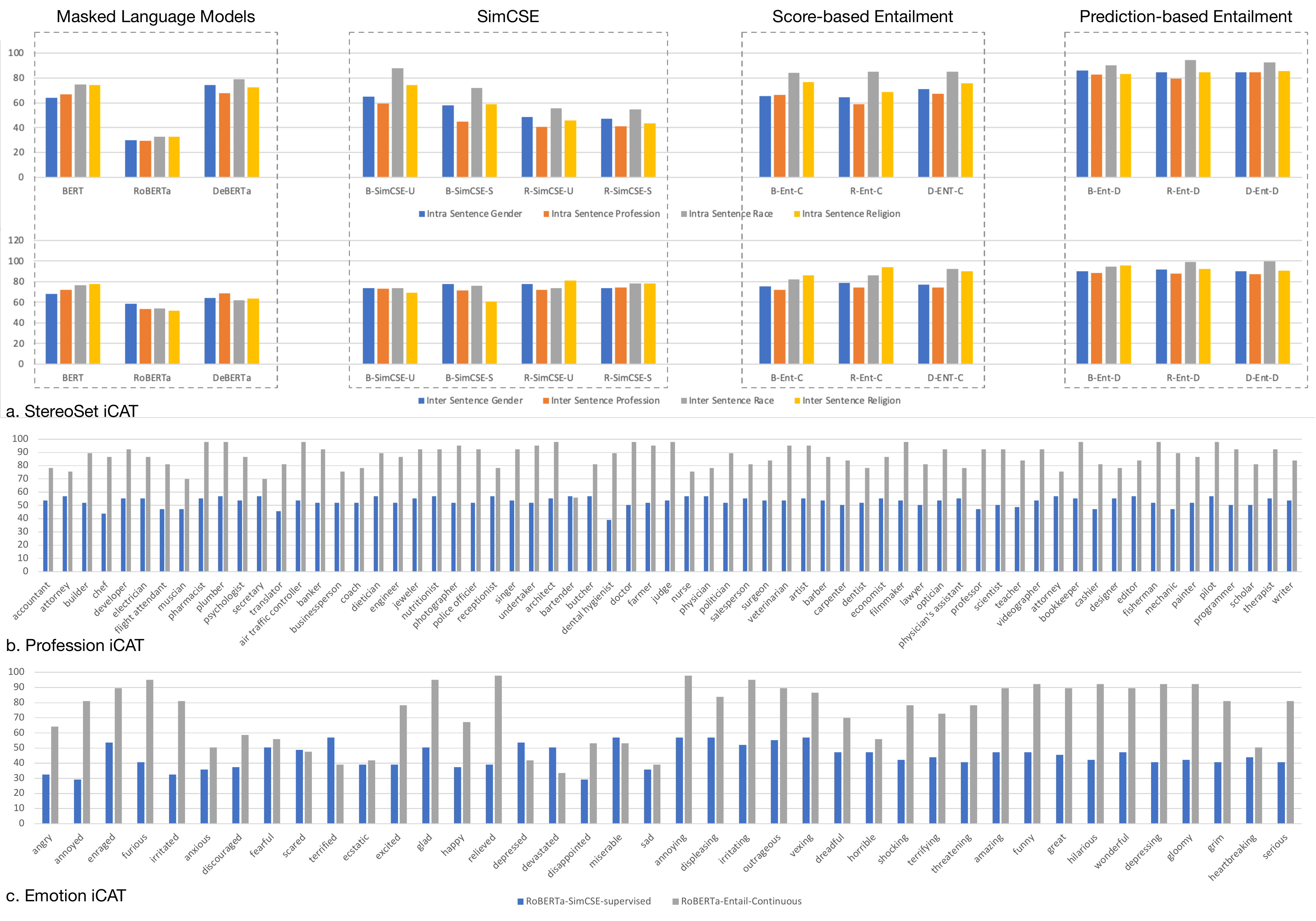}
\caption{Breakdown performance of pretrained and entailment language models on StereoSet, profession, and emotion bias tests. In StereoSet, we present the performance of all models. In the profession and emotion bias tests, we compare the performance of RoBERTa-SimCSE and the continuous RoBERTa entailment model.}
\label{fig:subtask}
\end{figure*}

For this task, all entailment models outperform all baseline models. The DeBERTa and RoBERTa models are significantly better than BERT-based models. For the continuous scoring models, the RoBERTa-based entailment model achieves the highest iCAT score (86.93\%), outperforming the best baseline model by 27\%. As for previous tasks, the discrete entailment scoring strategy is more fair. The best discrete entailment model, DeBERTa, achieves a high iCAT score (95.1\%), outperforming the best baseline model by 36\%. The exception is the RoBERTa-based entailment model. The continuous RoBERTa model outperforms the discrete model by almost 2\% iCAT score.

\noindent \textbf{Emotion bias test.} We use the emotion vocabulary sets, including 40 emotion state and situation words. We conduct context association tests on the gender-indicating nouns with the emotion words. On this task, the BERT and RoBERTa models have different behaviors. The RoBERTa-based SimCSE models outperform the pretrained RoBERTa model on both fairness and iCAT scores. However, the BERT SimCSE models are outperformed by the pretrained BERT model. The supervised RoBERTa model performs best among all baseline models, achieving 76\% iCAT score.

The entailment models outperform most baseline models. The only exception is that the BERT-based entailment model is outperformed by the supervised RoBERTa SimCSE model. However, the discrete entailment RoBERTa and DeBERTa entailment models outperform all baseline models by a large margin. The discrete RoBERTa entailment model outperforms the best baseline model by more than 11\%, and the DeBERTa entailment model outperforms the best baseline by 17\%.

\noindent \textbf{Summary.} We make the following observations:
\begin{itemize} \setlength{\itemsep}{0pt} \setlength{\parsep}{0pt}
\item SimCSE models achieve higher language model and gender recognition scores than pretrained models, but they are more biased.
\item The entailment models achieve significantly better performance than all baseline models in both language modeling and fairness metrics. The discrete scoring strategy is more fair than the continuous strategy, in general.
\end{itemize}
\begin{figure*}[t]
\centering
\includegraphics[width=\textwidth]{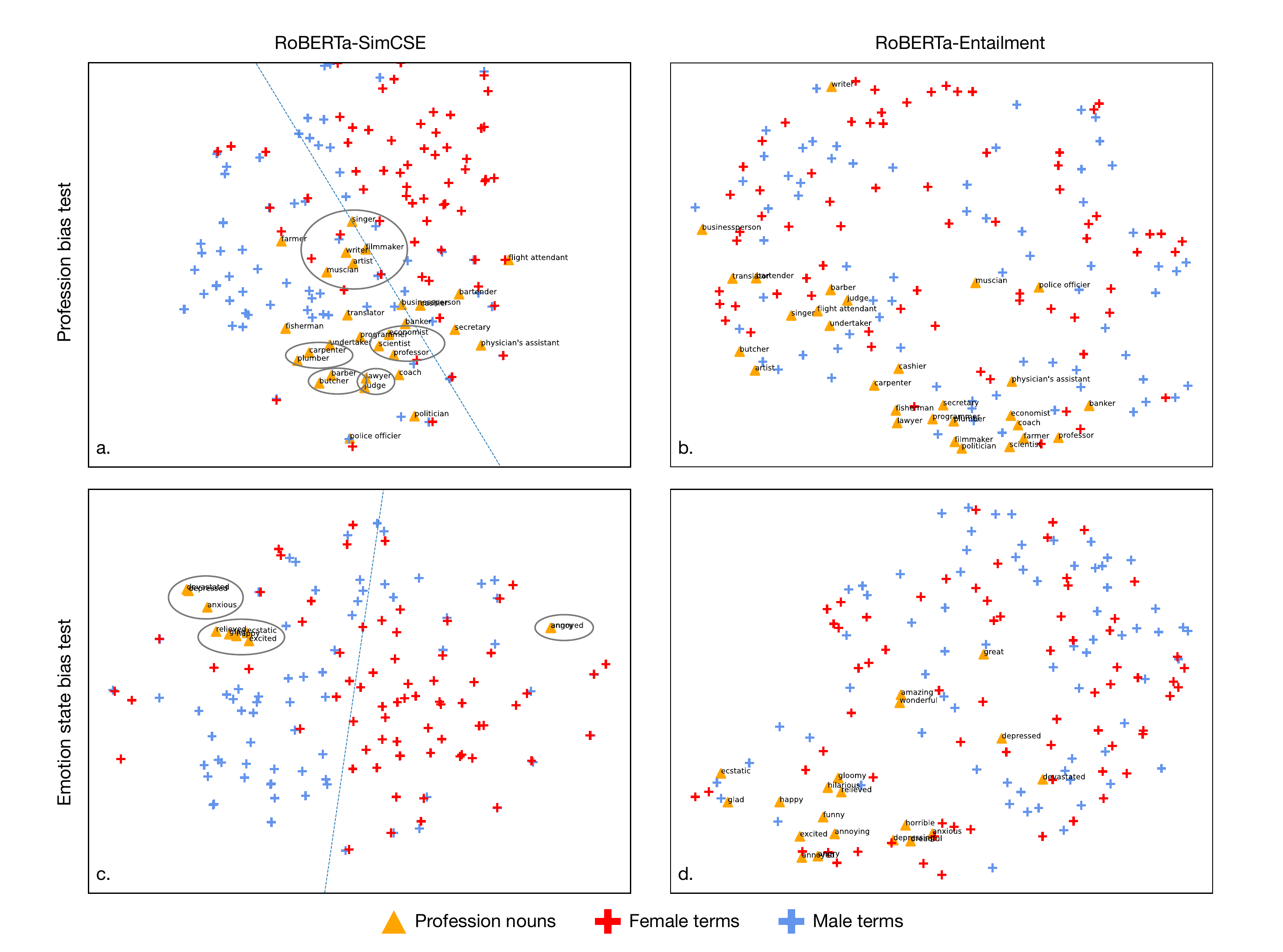}
\caption{The prompt analysis with RoBERTa-based SimCSE and entailment models on profession and emotion bias tests. In figures a. and c, different gender terms are separated by a boundary learned by a linear SVM, and the gray circles highlight correlated words. In Figure a., the circled clusters are \underline{[\textit{singer}, \textit{designer}, \textit{writer}, \textit{filmmaker}, \textit{artist}, \textit{musician}]}, \underline{[\textit{carpenter}, \textit{plumber}]}, \underline{[\textit{barber}, \textit{butcher}]}, \underline{[\textit{lawyer}, \textit{judge}]}, and \underline{[\textit{economist}, \textit{scientist}, \textit{professor}]}.
In Figure c., the circled clusters are \underline{[\textit{devastated}, \textit{depressed}, \textit{anxious}]}, \underline{[\textit{relieved}, \textit{ecstatic}, \textit{glad}, \textit{happy}, \textit{excited}]}, and \underline{[\textit{angry}, \textit{annoyed}]}.}
\label{fig:emb-analysis}
\end{figure*}

\section{Analysis}
\subsection{Performance Breakdown}
In the previous section, we reported the overall performance of each task. In this section, we analyze the performance of all sub-tasks. The StereoSet corpus has four sub-tasks, including gender, religion, profession, and race. The profession bias task has 65 different profession nouns as sub-tasks, and similarly, the emotion bias task has 40 sub-tasks. We break down and analyze the performance of the sub-tasks to investigate if the models conduct biased reasoning on sub-tasks, but achieve high average fairness scores. 

\noindent \textbf{StereoSet.} The breakdown iCAT scores of StereoSet sub-tasks is shown in Figure \ref{fig:subtask}.a, including the four sub-tasks under the intra- and inter-sentence settings. We do not find the entailment models to be biased on some of the sub-tasks. Instead, the entailment models consistently outperform the baseline pretrained models. We also note that the pretrained models based on different architectures achieve varying results on different tasks. In contrast,  the entailment model based on different architectures achieve stable iCAT scores. We also notice that the entailment models perform better on race and religion tasks. As shown in Table \ref{tab:results},  the performance of the discrete scoring models achieve better and more stable iCAT scores.

\noindent \textbf{Profession bias test.} We compare the breakdown performance of RoBERTa-based entailment and SimCSE models. As shown in Figure \ref{fig:subtask}.b, the iCAT scores on most profession terms of the entailment model outperforms the SimCSE model by more than 20\%. The only exception where the pretrained model outperforms the entailment models is the word ``\textit{Bartender}.'' The most significant improvement we achieved is almost 50\% iCAT score on the term ``\textit{dental hygienist}.''

\noindent \textbf{Emotion bias test.} We also test the RoBERTa-based models on different emotion state and situation terms. In all 40 emotion words, the entailment model outperforms the SimCSE model in 35 sub-tasks. The most biased emotion word of SimCSE is ``\textit{disappointed,}'' which is improved using the entailment model. On the other hand, the most biased emotion word of the entailment model is ``\textit{devastated}.'' Both models are relatively biased on the word ``sad,'' achieving lower than 40\% iCAT scores. The most significant improvement is on the word ``relieved.'' The sub-tasks that the entailment model does not outperform the pretrained models are ``\textit{scared},'' ``\textit{terrified},'' ``\textit{depressed},'' ``\textit{devastated},'' and ``\textit{miserable}.''


\subsection{Prompt Embedding Analysis}
We have found that the language modeling and fairness performance of entailment models are significantly higher than pretrained language models. In this section, we attempt to explain this phenomenon. To understand the difference between the entailment and pretrained models, we analyze the embedding of the gender terms and profession and emotion nouns. The results of the RoBERTa-based SimCSE and entailment models are shown in Figure \ref{fig:emb-analysis} with t-SNE \cite{van2008visualizing}.

The profession bias test results on RoBERTa-SimCSE is shown in Figure \ref{fig:emb-analysis}.a. We find that because of the strong representation ability of SimCSE, the embeddings of the profession and gender terms reflect the word similarities that aligns with human intuition. The boundary of the gender terms is detected by a linear SVM model \cite{hearst1998support,pedregosa2011scikit}. We find that the learned boundary separates terms of different genders with high accuracy. In addition, we notice that related profession terms group closely, as shown in the circles in Figure \ref{fig:emb-analysis}.a. In contrast, the word embedding distribution produced by the entailment model shown Figure \ref{fig:emb-analysis}.b appears to be more random. A similar phenomenon is observed on the emotion bias test. In Figure \ref{fig:emb-analysis}.c, nouns representing different genders are well-separated, and related words cluster closely. However in Figure \ref{fig:emb-analysis}.d, similar words are less correlated based on the entailment prompt embeddings.

The experimental results of both tasks and models indicate that the prompt embeddings learned by the entailment models contribute to logical reasoning rather than word coherence representation. Considering the fact that the entailment models perform significantly better than the pretrained models, we conclude that the biases are caused by the similarity-based learning objectives because such algorithms learn and reflect the biases in the training corpora. However, the textual entailment models learn logic without preserving textual similarities, leading to fairer performance.

\section{Conclusion}
In this work, we found that textual entailment learning reduces the bias of pretrained language models for sentence representation. We evaluated BERT, RoBERTa, and DeBERTa-based pretrained, SimCSE, and entailment models on stereotype, profession, and emotion bias tests. The textual entailment models outperform other models with significantly lower bias without other explicit debiasing processes, while preserving the language modeling ability, which results in significantly better idealized context association test scores. By analyzing the sentence embeddings, we found that the models relying on textual entailment produce less biased results by learning logic and reducing the amount of text coherence knowledge retained from the pretraining corpora containing existing social biases.

\section*{Acknowledgements}
We are grateful for the insightful comments and suggestions from the reviewers.

\section*{Ethics Statement}
We investigate the stereotypes and biases of pretrained language models and introduce the less biased textual entailment models that reduce bias on gender, profession, religion, and race. We noticed that the existing gender-related bias studies and corpora mainly focus on the binary gender setting, and we also follow this line of research because of data limitations. While such data limitation might disappoint a number of communities, we will extend this work to non-binary settings in future work.

\section*{Limitations}
As we described in the previous section, we studied the stereotypes including gender biases. However, we investigated under the binary gender setting, because of the limitation of the existing benchmarks. Furthermore, we evaluated medium-sized language models with around 350M parameters, but have not tested the largest language models yet. We only analyze the predictive bias on a set of gender-indicating vocabulary, but do not look into every example and explain the source of the learned bias in the pretraining corpora or social traditions.

On the other hand, there are further limitations in the benchmarks we study in this work, as pointed out by \citet{blodgett-etal-2021-stereotyping} that StereoSet is not perfect. On the other hand, some words in the vocabulary collected by \cite{lu2020gender} are rarely used, for example, ``poetess'' and ``manageress''. In future work, we will explore building more inclusive and comprehensive benchmarks to mitigate the limitations.

\bibliography{custom}

\begin{thebibliography}{49}
\expandafter\ifx\csname natexlab\endcsname\relax\def\natexlab#1{#1}\fi

\bibitem[{Bao and Qiao(2019)}]{bao2019transfer}
Xingce Bao and Qianqian Qiao. 2019.
\newblock Transfer learning from pre-trained bert for pronoun resolution.
\newblock In \emph{Proceedings of the first workshop on gender bias in natural
  language processing}, pages 82--88.

\bibitem[{Barikeri et~al.(2021)Barikeri, Lauscher, Vuli{\'c}, and
  Glava{\v{s}}}]{barikeri2021redditbias}
Soumya Barikeri, Anne Lauscher, Ivan Vuli{\'c}, and Goran Glava{\v{s}}. 2021.
\newblock Redditbias: A real-world resource for bias evaluation and debiasing
  of conversational language models.
\newblock \emph{arXiv preprint arXiv:2106.03521}.

\bibitem[{Bartl et~al.(2020)Bartl, Nissim, and Gatt}]{bartl2020unmasking}
Marion Bartl, Malvina Nissim, and Albert Gatt. 2020.
\newblock Unmasking contextual stereotypes: Measuring and mitigating bert's
  gender bias.
\newblock \emph{arXiv preprint arXiv:2010.14534}.

\bibitem[{Blodgett et~al.(2021)Blodgett, Lopez, Olteanu, Sim, and
  Wallach}]{blodgett-etal-2021-stereotyping}
Su~Lin Blodgett, Gilsinia Lopez, Alexandra Olteanu, Robert Sim, and Hanna
  Wallach. 2021.
\newblock \href {https://doi.org/10.18653/v1/2021.acl-long.81} {Stereotyping
  {N}orwegian salmon: An inventory of pitfalls in fairness benchmark datasets}.
\newblock In \emph{Proceedings of the 59th Annual Meeting of the Association
  for Computational Linguistics and the 11th International Joint Conference on
  Natural Language Processing (Volume 1: Long Papers)}, pages 1004--1015,
  Online. Association for Computational Linguistics.

\bibitem[{Bolukbasi et~al.(2016)Bolukbasi, Chang, Zou, Saligrama, and
  Kalai}]{bolukbasi2016man}
Tolga Bolukbasi, Kai-Wei Chang, James~Y Zou, Venkatesh Saligrama, and Adam~T
  Kalai. 2016.
\newblock Man is to computer programmer as woman is to homemaker? debiasing
  word embeddings.
\newblock \emph{Advances in neural information processing systems}, 29.

\bibitem[{Bordia and Bowman(2019)}]{bordia2019identifying}
Shikha Bordia and Samuel~R Bowman. 2019.
\newblock Identifying and reducing gender bias in word-level language models.
\newblock \emph{arXiv preprint arXiv:1904.03035}.

\bibitem[{Bowman et~al.(2015)Bowman, Angeli, Potts, and
  Manning}]{bowman2015large}
Samuel~R Bowman, Gabor Angeli, Christopher Potts, and Christopher~D Manning.
  2015.
\newblock A large annotated corpus for learning natural language inference.
\newblock \emph{arXiv preprint arXiv:1508.05326}.

\bibitem[{Brown et~al.(2020)Brown, Mann, Ryder, Subbiah, Kaplan, Dhariwal,
  Neelakantan, Shyam, Sastry, Askell et~al.}]{brown2020language}
Tom Brown, Benjamin Mann, Nick Ryder, Melanie Subbiah, Jared~D Kaplan, Prafulla
  Dhariwal, Arvind Neelakantan, Pranav Shyam, Girish Sastry, Amanda Askell,
  et~al. 2020.
\newblock Language models are few-shot learners.
\newblock \emph{Advances in neural information processing systems},
  33:1877--1901.

\bibitem[{Caliskan et~al.(2017)Caliskan, Bryson, and
  Narayanan}]{caliskan2017semantics}
Aylin Caliskan, Joanna~J Bryson, and Arvind Narayanan. 2017.
\newblock Semantics derived automatically from language corpora contain
  human-like biases.
\newblock \emph{Science}, 356(6334):183--186.

\bibitem[{Cao and Daum{\'e}~III(2019)}]{cao2019toward}
Yang~Trista Cao and Hal Daum{\'e}~III. 2019.
\newblock Toward gender-inclusive coreference resolution.
\newblock \emph{arXiv preprint arXiv:1910.13913}.

\bibitem[{Chaloner and Maldonado(2019)}]{chaloner2019measuring}
Kaytlin Chaloner and Alfredo Maldonado. 2019.
\newblock Measuring gender bias in word embeddings across domains and
  discovering new gender bias word categories.
\newblock In \emph{Proceedings of the First Workshop on Gender Bias in Natural
  Language Processing}, pages 25--32.

\bibitem[{Chuang et~al.(2022)Chuang, Dangovski, Luo, Zhang, Chang, Soljacic,
  Li, Yih, Kim, and Glass}]{chuang-etal-2022-diffcse}
Yung-Sung Chuang, Rumen Dangovski, Hongyin Luo, Yang Zhang, Shiyu Chang, Marin
  Soljacic, Shang-Wen Li, Scott Yih, Yoon Kim, and James Glass. 2022.
\newblock \href {https://doi.org/10.18653/v1/2022.naacl-main.311} {{D}iff{CSE}:
  Difference-based contrastive learning for sentence embeddings}.
\newblock In \emph{Proceedings of the 2022 Conference of the North American
  Chapter of the Association for Computational Linguistics: Human Language
  Technologies}, pages 4207--4218, Seattle, United States. Association for
  Computational Linguistics.

\bibitem[{Clark et~al.(2020)Clark, Luong, Le, and Manning}]{clark2020electra}
Kevin Clark, Minh-Thang Luong, Quoc~V Le, and Christopher~D Manning. 2020.
\newblock Electra: Pre-training text encoders as discriminators rather than
  generators.
\newblock \emph{arXiv preprint arXiv:2003.10555}.

\bibitem[{Costa-juss{\`a} et~al.(2019)Costa-juss{\`a}, Lin, and
  Espa{\~n}a-Bonet}]{costa2019gebiotoolkit}
Marta~R Costa-juss{\`a}, Pau~Li Lin, and Cristina Espa{\~n}a-Bonet. 2019.
\newblock Gebiotoolkit: Automatic extraction of gender-balanced multilingual
  corpus of wikipedia biographies.
\newblock \emph{arXiv preprint arXiv:1912.04778}.

\bibitem[{Dagan et~al.(2005)Dagan, Glickman, and Magnini}]{dagan2005pascal}
Ido Dagan, Oren Glickman, and Bernardo Magnini. 2005.
\newblock The pascal recognising textual entailment challenge.
\newblock In \emph{Machine learning challenges workshop}, pages 177--190.
  Springer.

\bibitem[{Devlin et~al.(2018)Devlin, Chang, Lee, and
  Toutanova}]{devlin2018bert}
Jacob Devlin, Ming-Wei Chang, Kenton Lee, and Kristina Toutanova. 2018.
\newblock Bert: Pre-training of deep bidirectional transformers for language
  understanding.
\newblock \emph{arXiv preprint arXiv:1810.04805}.

\bibitem[{Dinan et~al.(2019)Dinan, Fan, Williams, Urbanek, Kiela, and
  Weston}]{dinan2019queens}
Emily Dinan, Angela Fan, Adina Williams, Jack Urbanek, Douwe Kiela, and Jason
  Weston. 2019.
\newblock Queens are powerful too: Mitigating gender bias in dialogue
  generation.
\newblock \emph{arXiv preprint arXiv:1911.03842}.

\bibitem[{Elaraby et~al.(2018)Elaraby, Tawfik, Khaled, Hassan, and
  Osama}]{elaraby2018gender}
Mostafa Elaraby, Ahmed~Y Tawfik, Mahmoud Khaled, Hany Hassan, and Aly Osama.
  2018.
\newblock Gender aware spoken language translation applied to english-arabic.
\newblock In \emph{2018 2nd International Conference on Natural Language and
  Speech Processing (ICNLSP)}, pages 1--6. IEEE.

\bibitem[{Font and Costa-Jussa(2019)}]{font2019equalizing}
Joel~Escud{\'e} Font and Marta~R Costa-Jussa. 2019.
\newblock Equalizing gender biases in neural machine translation with word
  embeddings techniques.
\newblock \emph{arXiv preprint arXiv:1901.03116}.

\bibitem[{Gao et~al.(2020)Gao, Fisch, and Chen}]{gao2020making}
Tianyu Gao, Adam Fisch, and Danqi Chen. 2020.
\newblock Making pre-trained language models better -shot learners.
\newblock \emph{arXiv preprint arXiv:2012.15723}.

\bibitem[{Gao et~al.(2021)Gao, Yao, and Chen}]{gao-etal-2021-simcse}
Tianyu Gao, Xingcheng Yao, and Danqi Chen. 2021.
\newblock \href {https://doi.org/10.18653/v1/2021.emnlp-main.552} {{S}im{CSE}:
  Simple contrastive learning of sentence embeddings}.
\newblock In \emph{Proceedings of the 2021 Conference on Empirical Methods in
  Natural Language Processing}, pages 6894--6910, Online and Punta Cana,
  Dominican Republic. Association for Computational Linguistics.

\bibitem[{He et~al.(2020)He, Liu, Gao, and Chen}]{he2020deberta}
Pengcheng He, Xiaodong Liu, Jianfeng Gao, and Weizhu Chen. 2020.
\newblock Deberta: Decoding-enhanced bert with disentangled attention.
\newblock \emph{arXiv preprint arXiv:2006.03654}.

\bibitem[{Hearst et~al.(1998)Hearst, Dumais, Osuna, Platt, and
  Scholkopf}]{hearst1998support}
Marti~A. Hearst, Susan~T Dumais, Edgar Osuna, John Platt, and Bernhard
  Scholkopf. 1998.
\newblock Support vector machines.
\newblock \emph{IEEE Intelligent Systems and their applications}, 13(4):18--28.

\bibitem[{Kiritchenko and Mohammad(2018)}]{kiritchenko2018examining}
Svetlana Kiritchenko and Saif~M Mohammad. 2018.
\newblock Examining gender and race bias in two hundred sentiment analysis
  systems.
\newblock \emph{arXiv preprint arXiv:1805.04508}.

\bibitem[{Liang et~al.(2020)Liang, Li, Zheng, Lim, Salakhutdinov, and
  Morency}]{liang2020towards}
Paul~Pu Liang, Irene~Mengze Li, Emily Zheng, Yao~Chong Lim, Ruslan
  Salakhutdinov, and Louis-Philippe Morency. 2020.
\newblock Towards debiasing sentence representations.
\newblock \emph{arXiv preprint arXiv:2007.08100}.

\bibitem[{Liu et~al.(2019)Liu, Ott, Goyal, Du, Joshi, Chen, Levy, Lewis,
  Zettlemoyer, and Stoyanov}]{liu2019roberta}
Yinhan Liu, Myle Ott, Naman Goyal, Jingfei Du, Mandar Joshi, Danqi Chen, Omer
  Levy, Mike Lewis, Luke Zettlemoyer, and Veselin Stoyanov. 2019.
\newblock Roberta: A robustly optimized bert pretraining approach.
\newblock \emph{arXiv preprint arXiv:1907.11692}.

\bibitem[{Lu et~al.(2020)Lu, Mardziel, Wu, Amancharla, and
  Datta}]{lu2020gender}
Kaiji Lu, Piotr Mardziel, Fangjing Wu, Preetam Amancharla, and Anupam Datta.
  2020.
\newblock Gender bias in neural natural language processing.
\newblock In \emph{Logic, Language, and Security}, pages 189--202. Springer.

\bibitem[{Manzini et~al.(2019)Manzini, Lim, Tsvetkov, and
  Black}]{manzini2019black}
Thomas Manzini, Yao~Chong Lim, Yulia Tsvetkov, and Alan~W Black. 2019.
\newblock Black is to criminal as caucasian is to police: Detecting and
  removing multiclass bias in word embeddings.
\newblock \emph{arXiv preprint arXiv:1904.04047}.

\bibitem[{May et~al.(2019)May, Wang, Bordia, Bowman, and
  Rudinger}]{may2019measuring}
Chandler May, Alex Wang, Shikha Bordia, Samuel~R Bowman, and Rachel Rudinger.
  2019.
\newblock On measuring social biases in sentence encoders.
\newblock \emph{arXiv preprint arXiv:1903.10561}.

\bibitem[{Nadeem et~al.(2020)Nadeem, Bethke, and Reddy}]{nadeem2020stereoset}
Moin Nadeem, Anna Bethke, and Siva Reddy. 2020.
\newblock Stereoset: Measuring stereotypical bias in pretrained language
  models.
\newblock \emph{arXiv preprint arXiv:2004.09456}.

\bibitem[{Nangia et~al.(2020)Nangia, Vania, Bhalerao, and
  Bowman}]{nangia2020crows}
Nikita Nangia, Clara Vania, Rasika Bhalerao, and Samuel~R Bowman. 2020.
\newblock Crows-pairs: A challenge dataset for measuring social biases in
  masked language models.
\newblock \emph{arXiv preprint arXiv:2010.00133}.

\bibitem[{Obamuyide and Vlachos(2018)}]{obamuyide2018zero}
Abiola Obamuyide and Andreas Vlachos. 2018.
\newblock Zero-shot relation classification as textual entailment.
\newblock \emph{EMNLP 2018}, page~72.

\bibitem[{Pedregosa et~al.(2011)Pedregosa, Varoquaux, Gramfort, Michel,
  Thirion, Grisel, Blondel, Prettenhofer, Weiss, Dubourg
  et~al.}]{pedregosa2011scikit}
Fabian Pedregosa, Ga{\"e}l Varoquaux, Alexandre Gramfort, Vincent Michel,
  Bertrand Thirion, Olivier Grisel, Mathieu Blondel, Peter Prettenhofer, Ron
  Weiss, Vincent Dubourg, et~al. 2011.
\newblock Scikit-learn: Machine learning in python.
\newblock \emph{the Journal of machine Learning research}, 12:2825--2830.

\bibitem[{Raffel et~al.(2020)Raffel, Shazeer, Roberts, Lee, Narang, Matena,
  Zhou, Li, Liu et~al.}]{raffel2020exploring}
Colin Raffel, Noam Shazeer, Adam Roberts, Katherine Lee, Sharan Narang, Michael
  Matena, Yanqi Zhou, Wei Li, Peter~J Liu, et~al. 2020.
\newblock Exploring the limits of transfer learning with a unified text-to-text
  transformer.
\newblock \emph{J. Mach. Learn. Res.}, 21(140):1--67.

\bibitem[{Ravfogel et~al.(2020)Ravfogel, Elazar, Gonen, Twiton, and
  Goldberg}]{ravfogel2020null}
Shauli Ravfogel, Yanai Elazar, Hila Gonen, Michael Twiton, and Yoav Goldberg.
  2020.
\newblock Null it out: Guarding protected attributes by iterative nullspace
  projection.
\newblock \emph{arXiv preprint arXiv:2004.07667}.

\bibitem[{Recasens et~al.(2013)Recasens, Danescu-Niculescu-Mizil, and
  Jurafsky}]{recasens2013linguistic}
Marta Recasens, Cristian Danescu-Niculescu-Mizil, and Dan Jurafsky. 2013.
\newblock Linguistic models for analyzing and detecting biased language.
\newblock In \emph{Proceedings of the 51st Annual Meeting of the Association
  for Computational Linguistics (Volume 1: Long Papers)}, pages 1650--1659.

\bibitem[{Reimers and Gurevych(2019)}]{reimers2019sentence}
Nils Reimers and Iryna Gurevych. 2019.
\newblock Sentence-bert: Sentence embeddings using siamese bert-networks.
\newblock \emph{arXiv preprint arXiv:1908.10084}.

\bibitem[{Schick et~al.(2021)Schick, Udupa, and Sch{\"u}tze}]{schick2021self}
Timo Schick, Sahana Udupa, and Hinrich Sch{\"u}tze. 2021.
\newblock Self-diagnosis and self-debiasing: A proposal for reducing
  corpus-based bias in nlp.
\newblock \emph{Transactions of the Association for Computational Linguistics},
  9:1408--1424.

\bibitem[{Stanczak and Augenstein(2021)}]{stanczak2021survey}
Karolina Stanczak and Isabelle Augenstein. 2021.
\newblock A survey on gender bias in natural language processing.
\newblock \emph{arXiv preprint arXiv:2112.14168}.

\bibitem[{Thoppilan et~al.(2022)Thoppilan, De~Freitas, Hall, Shazeer,
  Kulshreshtha, Cheng, Jin, Bos, Baker, Du et~al.}]{thoppilan2022lamda}
Romal Thoppilan, Daniel De~Freitas, Jamie Hall, Noam Shazeer, Apoorv
  Kulshreshtha, Heng-Tze Cheng, Alicia Jin, Taylor Bos, Leslie Baker, Yu~Du,
  et~al. 2022.
\newblock Lamda: Language models for dialog applications.
\newblock \emph{arXiv preprint arXiv:2201.08239}.

\bibitem[{Thorne and Vlachos(2018)}]{thorne2018automated}
James Thorne and Andreas Vlachos. 2018.
\newblock Automated fact checking: Task formulations, methods and future
  directions.
\newblock \emph{arXiv preprint arXiv:1806.07687}.

\bibitem[{Van~der Maaten and Hinton(2008)}]{van2008visualizing}
Laurens Van~der Maaten and Geoffrey Hinton. 2008.
\newblock Visualizing data using t-sne.
\newblock \emph{Journal of machine learning research}, 9(11).

\bibitem[{Wang et~al.(2018)Wang, Singh, Michael, Hill, Levy, and
  Bowman}]{wang2018glue}
Alex Wang, Amanpreet Singh, Julian Michael, Felix Hill, Omer Levy, and Samuel~R
  Bowman. 2018.
\newblock Glue: A multi-task benchmark and analysis platform for natural
  language understanding.
\newblock \emph{arXiv preprint arXiv:1804.07461}.

\bibitem[{Wang et~al.(2022)Wang, Wang, Luo, Tan, Qiu, Yang, Shi, Huang, and
  Gao}]{wang2022towards}
Jianing Wang, Chengyu Wang, Fuli Luo, Chuanqi Tan, Minghui Qiu, Fei Yang,
  Qiuhui Shi, Songfang Huang, and Ming Gao. 2022.
\newblock Towards unified prompt tuning for -shot text classification.
\newblock \emph{arXiv preprint arXiv:2205.05313}.

\bibitem[{Webster et~al.(2020)Webster, Wang, Tenney, Beutel, Pitler, Pavlick,
  Chen, Chi, and Petrov}]{webster2020measuring}
Kellie Webster, Xuezhi Wang, Ian Tenney, Alex Beutel, Emily Pitler, Ellie
  Pavlick, Jilin Chen, Ed~Chi, and Slav Petrov. 2020.
\newblock Measuring and reducing gendered correlations in pre-trained models.
\newblock \emph{arXiv preprint arXiv:2010.06032}.

\bibitem[{Wei et~al.(2021)Wei, Bosma, Zhao, Guu, Yu, Lester, Du, Dai, and
  Le}]{wei2021finetuned}
Jason Wei, Maarten Bosma, Vincent~Y Zhao, Kelvin Guu, Adams~Wei Yu, Brian
  Lester, Nan Du, Andrew~M Dai, and Quoc~V Le. 2021.
\newblock Finetuned language models are zero-shot learners.
\newblock \emph{arXiv preprint arXiv:2109.01652}.

\bibitem[{Williams et~al.(2018)Williams, Nangia, and Bowman}]{N18-1101}
Adina Williams, Nikita Nangia, and Samuel Bowman. 2018.
\newblock \href {http://aclweb.org/anthology/N18-1101} {A broad-coverage
  challenge corpus for sentence understanding through inference}.
\newblock In \emph{Proceedings of the 2018 Conference of the North American
  Chapter of the Association for Computational Linguistics: Human Language
  Technologies, Volume 1 (Long Papers)}, pages 1112--1122. Association for
  Computational Linguistics.

\bibitem[{Yin et~al.(2019)Yin, Hay, and Roth}]{yin-etal-2019-benchmarking}
Wenpeng Yin, Jamaal Hay, and Dan Roth. 2019.
\newblock \href {https://doi.org/10.18653/v1/D19-1404} {Benchmarking zero-shot
  text classification: Datasets, evaluation and entailment approach}.
\newblock In \emph{Proceedings of the 2019 Conference on Empirical Methods in
  Natural Language Processing and the 9th International Joint Conference on
  Natural Language Processing (EMNLP-IJCNLP)}, pages 3914--3923, Hong Kong,
  China. Association for Computational Linguistics.

\bibitem[{Zmigrod et~al.(2019)Zmigrod, Mielke, Wallach, and
  Cotterell}]{zmigrod2019counterfactual}
Ran Zmigrod, Sabrina~J Mielke, Hanna Wallach, and Ryan Cotterell. 2019.
\newblock Counterfactual data augmentation for mitigating gender stereotypes in
  languages with rich morphology.
\newblock \emph{arXiv preprint arXiv:1906.04571}.

\end{thebibliography}
\bibliographystyle{acl_natbib}

\appendix



\end{document}